\documentclass[letterpaper, 10 pt, conference]{ieeeconf}  

\IEEEoverridecommandlockouts                              

\usepackage{graphics} 
\usepackage{epsfig} 
\usepackage{mathptmx} 
\usepackage{times} 
\usepackage{amsmath} 
\usepackage{amssymb}  
\usepackage{enumerate}
\usepackage{comment}
\usepackage{tablefootnote}
\usepackage{array}
\usepackage{booktabs}
\usepackage{extarrows}
\usepackage{soul}
\usepackage{xcolor}
\usepackage{xparse}
\usepackage[belowskip=-0.00pt,aboveskip=0pt]{caption}
\usepackage{cite}
\usepackage{amsfonts}
\usepackage{algorithmic}
\usepackage{textcomp}
\usepackage{xcolor}
\usepackage{algorithm}
\usepackage{breqn}
\usepackage{subcaption}
\usepackage{footnote}
\usepackage[bottom]{footmisc}
\usepackage{threeparttable}
\usepackage{tabularx}
\usepackage{ragged2e}
\usepackage{hyperref}

\newcommand{\Crs}[1]{{\color{red}{#1}}} 
\newcommand{\myHFed}{\emph{H\textsuperscript{2}-Fed}}
\newcommand{\eqdef}{\xlongequal{\text{def}}}%

\newcolumntype{C}[1]{>{\centering\arraybackslash}p{#1}}
\newcolumntype{P}[1]{>{\centering\arraybackslash}p{#1}}
\newcolumntype{M}[1]{>{\centering\arraybackslash}m{#1}}
\newcolumntype{L}[1]{>{\raggedright\arraybackslash}p{#1}}
\newcolumntype{R}[1]{>{\raggedleft\arraybackslash}p{#1}}
\newcolumntype{J}[1]{>{\justifying\arraybackslash}p{#1}}

\title{\LARGE \bf
Federated Learning Framework \\Coping with Hierarchical Heterogeneity in Cooperative ITS
}

\author{Rui Song$^{1,2}$ ,  Liguo Zhou$^{2}$ , Venkatnarayanan Lakshminarasimhan$^{2}$ , Andreas Festag$^{1,3}$ and Alois Knoll$^{2}$
\thanks{This work was supported by the German Federal Ministry for Digital and Transport (BMVI) in the projects ``KIVI -- KI im Verkehr Ingolstadt'' and ''5GoIng – 5G Innovation Concept Ingolstadt''.}%
\thanks{$^{1}$Rui Song and Andreas Festag are with Fraunhofer Institute for Transportation and Infrastructure Systems IVI, Ingolstadt, Germany, e-mail:
        {\tt\small \{rui.song, andreas.festag\}@ivi.fraunhofer.de}.}%
\thanks{$^{2}$Rui Song, Liguo Zhou, Venkatnarayanan Lakshminarasimhan and Alois Knoll are with Technical University of Munich, Robotics, Artificial Intelligence and Real-Time Systems, Garching, Germany, e-mail:
        {\tt\small \{rui.song, liguo.zhou, venkat.lakshmi\}@tum.de, knoll@in.tum.de}.}%
\thanks{$^{3}$Andreas Festag is with Technische Hochschule Ingolstadt, CARISSMA Institute for Electric, COnnected, and Secure Mobility (C-ECOS), Ingolstadt, Germany, e-mail:
        {\tt\small andreas.festag@carissma.eu}.}%
}

\begin{document}

\maketitle
\thispagestyle{empty}
\pagestyle{empty}

\begin{abstract}

Deep learning is a key approach for the environment perception function of Cooperative Intelligent Transportation Systems (C-ITS) with autonomous vehicles and smart traffic infrastructure. %
The performance of the object recognition and detection strongly depends on the data volume in the model training, which is usually not sufficient due to the limited data collected by a typically small fleet of test vehicles. %
In today's C-ITS, smart traffic participants are capable of timely generating and transmitting a large amount of data. However, these data can not be used for model training directly due to privacy constraints. %
In this paper, we introduce a federated learning framework coping with Hierarchical Heterogeneity (\myHFed), which can notably enhance the conventional 
pre-trained deep learning model. The framework exploits data from connected public traffic agents in vehicular networks without affecting user data privacy. 
By coordinating existing traffic infrastructure, including roadside units and road traffic clouds, the model parameters are efficiently disseminated by vehicular communications and hierarchically aggregated.
Considering the individual heterogeneity of data distribution, computational and communication capabilities across traffic agents and roadside units, we employ a novel method that addresses the heterogeneity in different aggregation layers of the framework architecture, i.e., aggregation in layers of roadside units and cloud. 
The experimental results indicate that our method can well balance the learning accuracy and stability according to the knowledge of heterogeneity in current communication networks. Comparing to other baseline approaches, the evaluation on federated datasets shows that our framework is more general and capable especially in application scenarios with low communication quality. Even when 90\% of the agents are timely disconnected, the pre-trained deep learning model can still be forced to converge stably, and its accuracy can be enhanced from 68\% to over 90\% after convergence.  
\end{abstract}


\section{Introduction}
\label{sec:introduction}

\begin{figure*}[ht]
\centering
\includegraphics[width=0.9\linewidth]
{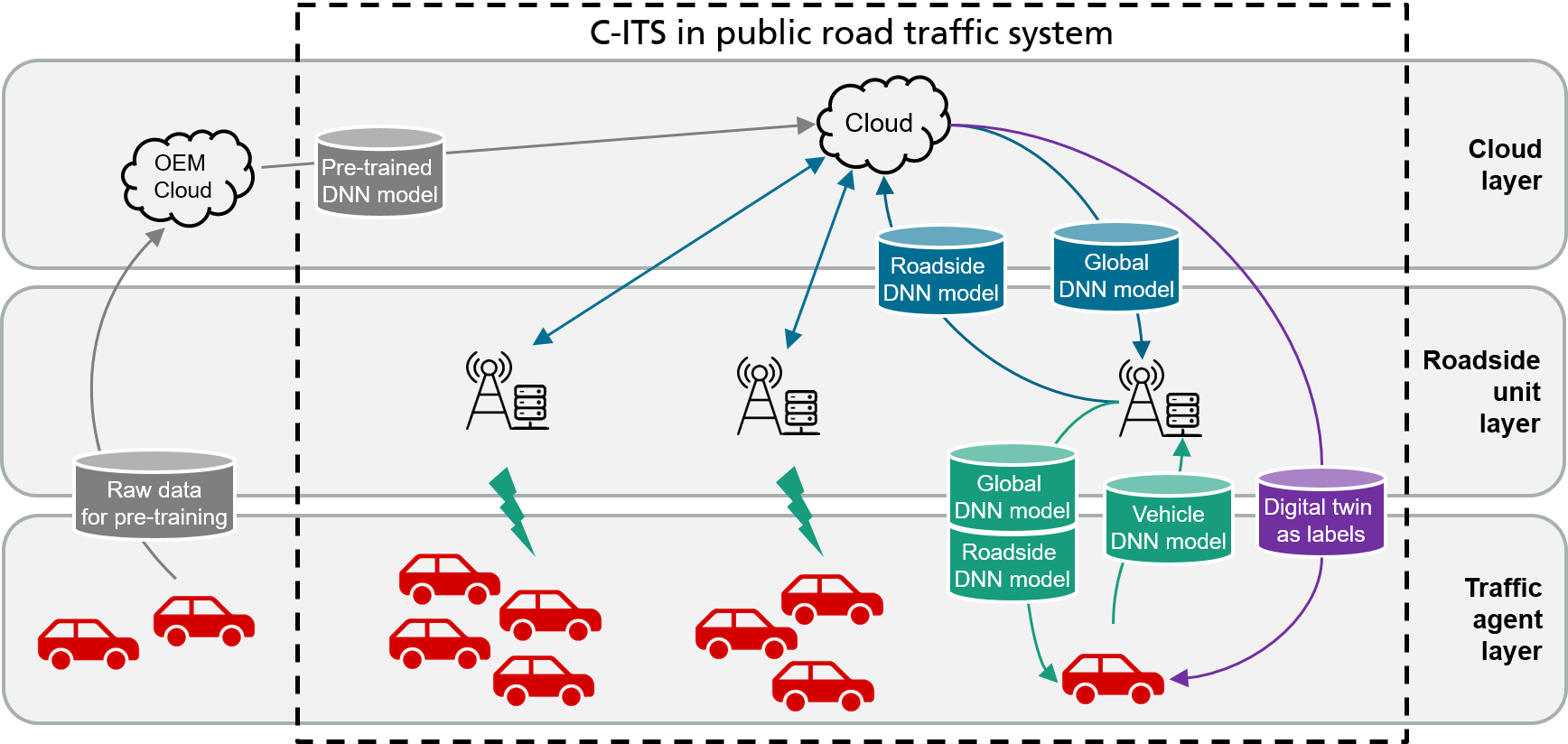}
\vspace{3mm}
\caption{Proposed framework for federated learning in a Cooperative Intelligent Transportation System (C-ITS): Blue, green and purple arrows indicate I2N, V2I and N2V communication, respectively. The gray line represents OEM-internal communication.}
\label{fig:sys}
\end{figure*}

Artificial intelligence, especially deep learning, 
is one of the most important technologies in today's \emph{Cooperative Intelligent Transportation System} (C-ITS) development. %
Its application requires a huge amount of data collected on traffic roads. However, the deep learning models, i.e., \emph{Deep Neural Network} (DNN) models, are usually trained by an insufficient amount of data because automotive manufacturers (Original Equipment Manufacturers, OEMs) train their DNN models for autonomous driving only with data from a very limited number of test vehicles. The resulting bias in the data distribution can lead to errors in autonomous driving functions and threaten traffic safety.

In most regions, the acquisition of raw data from public road traffic is challenged due to privacy constraints. Fortunately, federated learning~\cite{pmlr-v54-mcmahan17a} has been proposed as a distributed learning approach that does not affect the privacy of user data. %
It provides C-ITS a great opportunity to gain data from public road traffic and enhance the DNN models, which can be centralized pre-trained on internal or open datasets~\cite{cress2022a9}.  \emph{Vehicle-to-Everything} (V2X) communication~\cite{Festag-11p-to-5G}
enables the data exchange among vehicles, roadside infrastructure, backends and cloud systems, and facilitates federated learning in C-ITS. Based on abundant \emph{Roadside Units} (RSUs), digital twins~\cite{Cress2021_ITSReview} can provide accurate labels for the learning process.

As many other federated learning application scenarios, \mbox{C-ITS} faces heterogeneity problems. 
These problems exist due to the lack of full access and control to traffic agents. In C-ITS, heterogeneity derives mainly from \emph{(i)}~various available computational capabilities, 
\emph{(ii)} timely varied communication quality, and \emph{(iii)} individual local training datasets, known as not-independent and identically distributed (Non-IID) data. It can lead to inconsistency in the training process across the traffic agents and cause low convergence speed and instability in federated learning in C-ITS.

To guarantee the success of federated learning and enhance the pre-trained DNN models in C-ITS, we develop a novel federated learning framework as shown in Fig.~\ref{fig:sys}. The framework allows all connected vehicles to collaboratively train a common DNN model via hybrid vehicular communications, i.e., \emph{Vehicle-to-Infrastructure} (V2I) and \emph{Vehicle-to-Network} (V2N). Through sharing both global and roadside federated learning models in our framework, the effect of hierarchical heterogeneity on the learning process can be beneficially reduced, and hence the performance of pre-trained DNN models is powered by data acquired from public traffic.

\noindent
The key contributions of the paper are:

\begin{itemize}
    \item Formulate the problem with Non-IID data across RSUs, individual communication and computation capabilities at the connected traffic agents, and analyze the impact on the federated learning performance in vehicular networks,
    \item Address individual heterogeneity in different layers of the hierarchical federated learning system and develop multiple proximal terms in the learning algorithm,
    \item Propose a general and empirical federated forward training framework for enhancing deep learning tasks in C-ITS by means of V2X networks without affecting the data privacy, 
    \item Evaluate our methods by federated training a DNN model on benchmark datasets considering the Non-IID data properties and heterogeneous communication qualities in C-ITS,
    \item Provide comprehensive experiment results and in-depth analysis of parameters in our framework with heterogeneously connected traffic agents in public road traffic,
    \item Compare our framework with other federated learning approaches and address the trade-off between stability and accuracy in federated learning application scenarios. The software implementation of our framework is publicly available as open source at GitHub.\footnote{\url{https://github.com/rruisong/H2-Fed}}.
\end{itemize}

The remainder of this paper is structured as follows. Section~\ref{sec:related_work} summarizes related work in federated learning and its application in C-ITS. The problem with individual hierarchical heterogeneity in vehicular networks and federated learning is formulated in Section~\ref{sec:problem_formulation}. The federated learning methods are introduced in Section~\ref{sec:method} and the proposed {\myHFed} framework is presented in Section~\ref{sec:system}. The system is comprehensively evaluated in Section~\ref{sec:evaluation} considering heterogeneity in individual layers of the framework architecture with respect to C-ITS scenarios. Section~\ref{sec:conclusion} provides a summary and concludes.

\section{Related Work}
\label{sec:related_work}
\vspace{-1mm}
Federated learning is a particular distributed learning framework, which was proposed in~\cite{pmlr-v54-mcmahan17a} as \emph{FedAvg} and can be employed for training models without accessing the user's data in massively distributed systems. To achieve joint learning in a federated fashion, especially wireless communication networks for sharing the locally training results across infrastructure are required. Specifically, deployment of C-V2X networks~\cite{9128410, Festag2021,delooz2022analysis} according to corresponding standards~\cite{Festag-Commag, SongFISITA2021} facilities federated learning in C-ITS, where many traffic services and applications need to be developed based on DNN models, e.g., object classifier~\cite{ResNet}, quality monitors~\cite{zhou2021semml,song2022edge}, cooperative detectors~\cite{xu2021opv2v,xu2022v2xvit,chen2022model,zimmer2022real}, etc..
In this context, \cite{Posner2021_FL_V2X,FL_VehEdgeCloud,Elbir2020_FL_Veh} indicate the great potential benefits of federated learning in C-ITS.

As with many other federated learning application scenarios, heterogeneity problems exist while implementing federated as well. Research in~\cite{li2020federated, Elbir2020_FL_Veh, 9521822} and others address heterogeneity problems and provide general federated learning algorithms (e.g., \emph{FedProx}) by adjusting the training objective to reduce the influence of straggling devices. However, depending on specific federated learning applications and associated communication infrastructure, the heterogeneity can be caused by various 
reasons and needs to be optimized individually.


\begin{table*}[ht]
\begin{threeparttable}
\caption{\centering Overview of heterogeneity metrics while hierarchically applying federated learning system in C-ITS}
\label{table:comparison}
\begin{tabular}{M{3.2cm}M{1cm}M{4.8cm}M{7.1cm}}
   \toprule 
    \centering\textbf{Heterogeneity metric} & \centering\textbf{Notation} & \centering\textbf{Main cause}  & \textbf{Effect on Federated Learning}\\
    \midrule 
   \vspace{1mm}
    Local Aggregation Round  & LAR & Diverse Quality of Service (QoS) & The pre-aggregation\,\footnote{1} round will be done in each RSU \\
   \vspace{1mm}
    Connection Success Ratio  & CSR & Varied communication quality & The number of agents connected for federated learning\\
   \vspace{1mm}
    Stable Connection Duration & SCD & Varied communication quality & Stable participating time of one agent, once it is connected \\
    Full-task Success Ratio\,\footnote{2} & FSR & Individual available computation resource & Finished training epochs in each agent\\
    \bottomrule 
\end{tabular}
\small
\begin{tablenotes}
    \scriptsize 
    \item[1] Aggregation in an RSU before the global aggregation in the cloud.
    \item[2] FSR is not the focus in this paper and has 
    a similar effect as CSR.
\end{tablenotes}
\end{threeparttable}
\end{table*}

Moreover, to utilize the hierarchical infrastructure for federated learning in C-ITS, multi-layer aggregation can be used. 
~\cite{9207469} provides a federated learning framework in hierarchical clustering steps and~\cite{9148862} proposes hierarchical federated learning (\emph{HierFAVG}) in a client-edge-cloud system, which can be regarded as similar to C-ITS considered in this paper. However, heterogeneity is not considered. 
\cite{9054634} focuses on federated learning in cellular networks but do not take the possible direct communication among vehicles and with RSUs via sidelink into account.  
Sidelink is regarded as a typical communication method in V2X networks and is considerably faster than up- and downlink~\cite{Hegde2020-VNC,5GNRV2X-tutorial}. 
Meanwhile, due to the various V2X messaging services, especially for traffic safety, as evaluated in~\cite{Kuehlmorgen2020}, the priority of data transmission for federated learning in V2X networks can be low compared to safety-related messages, which causes additional time-variant heterogeneity problems across traffic agents.

In this paper, we develop a hierarchical federated learning framework specifically for hybrid V2X networks in C-ITS with consideration of heterogeneity problems. Differentiating from other existing research work, we analyze the heterogeneity caused by V2X communication and available computational capabilities in traffic agents. By employing multiple proximal terms in the federated learning algorithm, the individual heterogeneity in different layers of the framework can be coped for the successes of model training.

\section{Heterogeneity Problems}
\label{sec:problem_formulation}

\noindent  \textbf{Heterogeneity in datasets.} 
We assume that RSUs are equipped with homogeneous computational capabilities and benefit from a stable communication quality. Particularly, the traffic cloud reliably receives the aggregated models from all RSUs. However, the models generated in each RSU are based on a dataset that is notably different from other RSUs due to the individual position of each RSU. For instance, the model in an RSU equipped at a highway is trained with data in highway scenarios, where the involved dynamic objects in the datasets are typically passenger cars or trucks that move fast in simple road typologies. In contrast, there are various types of dynamic objects with low speed in complex road networks in the datasets, when an RSU is located in cities. In addition, even urban scenarios in various locations can be marked distinct, e.g., at crossings, roundabouts, etc. Therefore, we believe that the models generated across RSUs are based on Non-IID datasets. Besides, the diverse average traffic flows cause the unbalanced agent number at RSUs, which can finally impact the learning results while aggregating the models at the traffic cloud. 

\noindent  \textbf{Heterogeneity in communication.} 
The direct communication between traffic agents and RSUs achieves local aggregation (or pre-aggregation) in federated learning at each RSU, where traffic agents and the RSU represent the clients and server in non-hierarchical federated learning. Then, before one global aggregation, each RSU can implement several \emph{Local Aggregation Rounds} ($LAR$), which accelerates the learning process by means of pre-aggregation between two global aggregation steps. Through sidelink communication in Cellular-V2X, 
the models at the RSU can be pre-aggregated up to 50 times, if we aggregate models from RSUs every $1$\,s globally in the cloud. In fact, due to the various other traffic services in V2X communication, we need to set the $LAR$ in a proper range with respect to the current \emph{Quality of Service} (QoS) at each RSU. Additionally, according to the current data traffic in the communication channel, not all traffic agents can continuously upload their models to the RSUs. Some traffic agents might need to allocate the communication resource for other tasks with higher priority. The metric \emph{Connection Success Ratio} ($CSR$) indicates the current connection situation specifically for federated learning at each RSU $j$:
\begin{equation}
    \label{Eq:measurement}
    CSR_j = \frac{N_{j,{connected\: agent}}}{N_{j,{participant}}}, (CSR \in [0,1]),
\end{equation}
where $N_{j,{participant}}$ is the number of all connected agents at RSU $j$ and $N_{j,{connected\:agent}}$ is the number of successfully connected agents, which stably upload the models in a predefined duration, specified as \emph{Stable Connection Duration} (SCD) in seconds.

\noindent \textbf{Heterogeneity in computation.} 
Apart from communication, the available computation resources can also vary in each traffic agent. This leads to the fact that not all traffic agents finish the expected local training epochs $E$. Normally, federated learning requires each agent to finish local training by the same $E$. However, the different computational capabilities and currently occupied computational resources cannot always satisfy the request on $E$ within a specific aggregation frequency, which causes further heterogeneity. 
\emph{Full-task Success Ratio} (FSR) indicates the ratio of traffic agents that can finish the requested $E$ in time. Similarly, $FSR$ can also be individual in each FSR. Note that if an agent cannot even finish one epoch, its results will be discarded, which has the same effect as a failed connection. Thus, in this paper, we will address heterogeneity especially considering varied $CSR$. All heterogeneity metrics are summarized in Tab.~\ref{table:comparison}

\section{Federated Learning Methods}
\label{sec:method}

Federated learning can enhance the pre-trained DNN model for intelligent traffic applications since it can further improve the model with broad data collected by public traffic without affecting user data privacy. Considering the V2X communication networks for federated learning, the DNN models from agents can be uploaded via direct communication. Thanks to the high-defined sensors at intelligent infrastructure, the highly accurate perception are achieved and can be set for 3D digital twin services, which can be deduced into sensor raw data from a global coordinate system and taken as labels for supervised federated learning implementation.

As a baseline method, we first employ FedAvg~\cite{pmlr-v54-mcmahan17a} without dropping any clients in the hierarchical federated learning framework. For an RSU $k$, the aim is to minimize: 

\begin{equation}
    \label{Eq:rsu-fedavg}
    \min_{w} f_k(w) = \sum_{i \in \mathcal{P}_k^t} \frac{n_{i,k}}{n_k} F_{i,k}(w), 
\end{equation}
where $\mathcal{P}_k^t$ is a partition of traffic agents selected at time point $t$ at RSU $k$ and $n_k$ is the total number of data points in all traffic agents selected by the RSU federated learning orchestrator. The agent $i$ trains with $n_{i,k}$ data points and $F_{i,k}(w)$ is averaged optimized objective in the agent $i$ with respect to non-convex DNN optimization.
If $K$ RSU train a DNN model jointly, we can extend the objective form in (\ref{Eq:rsu-fedavg}) as:

\begin{equation}
    \label{Eq:rsu-fedavg-ext}
    \min_{w} h(w) = \sum_{k=1}^{K} \sum_{i \in \mathcal{P}_k^t} \frac{n_{i,k}}{n_k} F_{i,k}(w) 
\end{equation}

To address the individual heterogeneity in each layer of the federated learning system, as analyzed in Sec.~\ref{sec:problem_formulation}, we employ the $\gamma_k^t$-inexact solution~\cite{li2020federated} with multiple proximal terms in the loss function of the hierarchical federated learning. The  aim of the traffic agent $i$ connecting RSU $k$ is then to minimize:

\begin{equation}
    \label{Eq:prox-hfl}
    \min_{w_{i,k}} h_k(\mathord{\cdot}) = F_k(w_{i,k}) + \sum_{l=1}^{L} \frac{\mu_{k,l}}{2} ||w_{i,k}-w_l||^2
\end{equation}
where $L$ is the number of aggregation layers. We note $\mathcal{M} = \{\mu_{k,l}|k=1,2,...,K, l=1,2,...,L\}$ as a set of main parameters in \ref{Eq:prox-hfl}. In vehicular networks, $L\,=\,2$. When $l\,=\,1$, $\mu_1$ indicates the heterogeneity of data from traffic agents and $w_1$ is the current weight of DNN model in the connected RSU. For $l\,=\,2$, $\mu_2$ indicates the heterogeneity of models across RSUs and $w_2$ is the current model weight in the traffic cloud.\\
To simplify the notation, we define 
\begin{equation}
    \label{Eq:wdef}
    \begin{split}
        &\textit{for an RSU $k$: }  w_k \eqdef w_{l\,=\,1}, \\
        &\textit{for the traffic cloud: }  w \eqdef w_{l\,=\,2}.
    \end{split}
\end{equation}
We can rewrite (\ref{Eq:prox-hfl}) as a double-layer aggregation system in accordance with the application of hierarchical federated learning in vehicular networks as
\begin{equation}
    \label{Eq:HtHFed}
    \min_{w_{i,k}} h_k(\mathord{\cdot}) = F_k(w_{i,k}) + {\frac{\mu_{k,1}}{2} ||w_{i,k}-w_k||^2} \\ 
    + {\frac{\mu_{k,2}}{2} ||w_{i,k}-w||^2},
\end{equation}
where for each RSU $k$, the set of main parameters is $\mathcal{M}_k = \{\mu_{k,l}|l=1,2\}$.

The method defined in (\ref{Eq:HtHFed}) reduces the effect of individual heterogeneity of data in different layers by setting the parameters $\mu_{k,1}$ and $\mu_{k,2}$ in $\mathcal{M}_k$, which can stabilize federated optimization through adding the additional trailers in the loss. Though the multiple proximal terms compared with a single term might slow down the convergence of training~\cite{li2020federated}, it provides more tunable parameters to separately consider the heterogeneity among RSUs and across agents in one RSU, which will be evaluated in Section~\ref{sec:evaluation}.

\section{System Framework}
\label{sec:system}

We divide the C-ITS into three layers: traffic agent layer, roadside unit layer, and cloud layer (Fig.~\ref{fig:sys}). We first consider a pre-trained process, where raw data are collected by automotive OEMs internal test vehicles the in traffic agent layer. A DNN model is trained in a centralized fashion based on the data managed in the cloud layer. Since the amount of the collected data is not sufficient for training, the pre-trained DNN model is sent to a public traffic cloud, which can orchestrate federated learning for model enhancements in public C-ITS.

The cloud shares the pre-trained DNN model with all connected RSUs in the roadside unit layer. This pre-trained DNN model is taken as the initial global and roadside FL model and forwarded to the traffic agents by the RSUs. Once the traffic agents in the traffic agent layer receive the duplicated model via V2I communication, the training process is run based on the cached datasets. The data labeling can be achieved by digital twin services via V2N communication~\cite{Cress2021_ITSReview}.

According to the specific federated learning orchestration, the traffic agents train the DNN model several epochs and share it back to their connected RSUs. Then RSUs aggregate the models from the selected partition of agents and update the roadside FL model. Before uploading the model to the cloud, the local aggregation in the roadside unit layer can be implemented a few times, as the model exchange between the roadside unit and traffic agent layers using sidelink communication is considered to be faster than up- and downlink. The roadside unit layer can be regarded as the first aggregation layer. After a necessary pre-aggregation, the roadside FL models from all RSUs are sent to the cloud via I2N communication and aggregated globally. The cloud layer is then the second aggregation layer. The new global FL model is sent back to all RSUs, where the old roadside and global FL models in the roadside unit layer are replaced.

To address the heterogeneity problems in C-ITS as analyzed in the previous Section~\ref{sec:problem_formulation}, we employ the method defined in~(\ref{Eq:HtHFed}) for federated optimization in hierarchical systems and propose a \emph{Federated Learning Framework Coping with Hierarchical Heterogeneity} ({\myHFed}). The pseudo-code for traffic agents, RSUs and the traffic cloud are given in Algorithm~\ref{alg:agent},~\ref{alg:rsu} and~\ref{alg:cloud}, respectively. By designing parameters in the traffic agent and RSU layers, our {\myHFed} system can be deployed as a general framework for other federated learning methods with respect to (\ref{Eq:prox-hfl}) : \emph{(i)} $\mu_{k,l}\,=\,0$ and $L\,=\,1$ correspond to the idea of \emph{FedAvg}~\cite{pmlr-v54-mcmahan17a}; \emph{(ii)} $\mu_{k,l}\,>\,0$ and $L\,=\,1$ are equivalent to \emph{FedProx}~\cite{li2020federated}; \emph{(iii)} $\mu_{k,l}\,=\,0$ and $L\,>\,1$ match \emph{HierFAVG}~\cite{9148862}. Furthermore, the trade-off between stability and accuracy can be tuned by designing all parameters for traffic agents at different RSUs in {\myHFed}. This will be evaluated in Sec.~\ref{sec:evaluation}.

\begin{algorithm}[t]
 \caption{\raggedright Traffic agent $i$ connecting RSU $k$ in {\myHFed}}
 \label{alg:agent}
 \begin{algorithmic}[1]
 \renewcommand{\algorithmicrequire}{\textbf{Input:}}
 \renewcommand{\algorithmicensure}{\textbf{Output:}}
 \newcommand{\algorithmicbreak}{\textbf{break}}
 \newcommand{\BREAK}{\STATE \algorithmicbreak}
 
 \REQUIRE {$\eta$, $\mathcal{M}_{i,k}$, $E$, digital twin data}
 \ENSURE  {$w_{i,k}$}
    \STATE receive $w_k$ and $w$ from RSU $k$
    \STATE generate labels from digital twin data
    \FOR{each epoch $\tau$ = 1,...,E}
        \STATE $w_{i,k} \gets \eta\nabla \mathcal{L}^*_{\mathcal{M}_{i,k}}(w_{i,k};w_{k};w)$ based on Equation (\ref{Eq:HtHFed})
    \ENDFOR
    \STATE send $w_{i,k}$ to RSU $k$
\end{algorithmic}
\end{algorithm}

\begin{algorithm}[t]
 \caption{\raggedright RSU $k$ in {\myHFed}}
 \label{alg:rsu}
 \begin{algorithmic}[1]
 \renewcommand{\algorithmicrequire}{\textbf{Input:}}
 \renewcommand{\algorithmicensure}{\textbf{Output:}}
 \newcommand{\algorithmicbreak}{\textbf{break}}
 \newcommand{\BREAK}{\STATE \algorithmicbreak}
 \REQUIRE {$LAR$}
 \ENSURE  {$w_{i,k}$}
     \FOR{each local round}
        \STATE download $w$ from cloud
        \STATE generate $\mathcal{P}_k^t$ based on orchestration
        \FOR{each clients $i \in \mathcal{P}_k^t$}
            \STATE send $w$ and $w_k$ to client $i$
            \STATE receive $w_{i,k}$ from client $i$
        \ENDFOR
        \STATE $w_k \gets \sum_{i\in{\mathcal{P}_k^t}} \frac{n_{i,k}}{n_k}{w_k}$
       \ENDFOR 
    \STATE upload $w_{k}$ to cloud
\end{algorithmic}
\end{algorithm}

\begin{algorithm}[t]
 \caption{\raggedright Cloud in {\myHFed}}
 \label{alg:cloud}
 \begin{algorithmic}[1]
 \renewcommand{\algorithmicrequire}{\textbf{Input:}}
 \renewcommand{\algorithmicensure}{\textbf{Output:}}
 \newcommand{\algorithmicbreak}{\textbf{break}}
 \newcommand{\BREAK}{\STATE \algorithmicbreak}
 \REQUIRE {initial $w$ from a private cloud}
 \ENSURE  {$w$}
     \FOR{each global round}
        \FOR{each RSU $k$}
            \STATE send $w$ to RSU $k$
            \STATE receive $w_k$ from RSU $k$
        \ENDFOR 
     \STATE $w \gets \sum_k \frac{n_k}{n}{w_k}$
     \ENDFOR 
\end{algorithmic}
\end{algorithm}

\section{Experimental results}
\label{sec:evaluation}
\begin{figure*}[htb]
\includegraphics[trim=0 0 0 0,clip,width=0.98\linewidth]{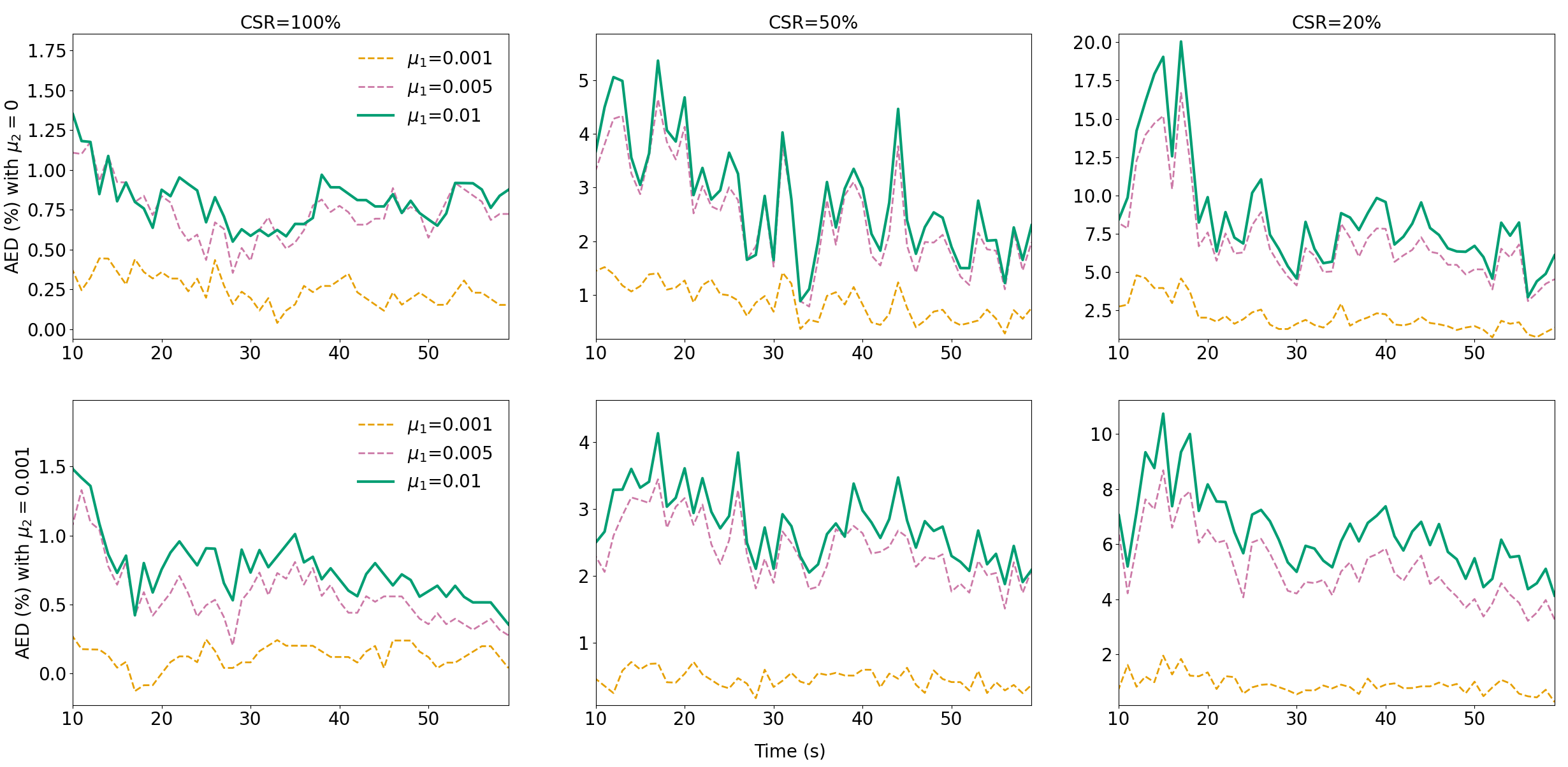}
\caption{Adjusting $\mu_1$ raises the \emph{Accuracy Enhancement Degree} (AED) in scenarios with heterogeneous communication quality.}
\label{fig:Mue1}
\end{figure*}

\begin{figure*}[ht]
\includegraphics[trim=0 0 0 0,clip,width=0.97\linewidth]{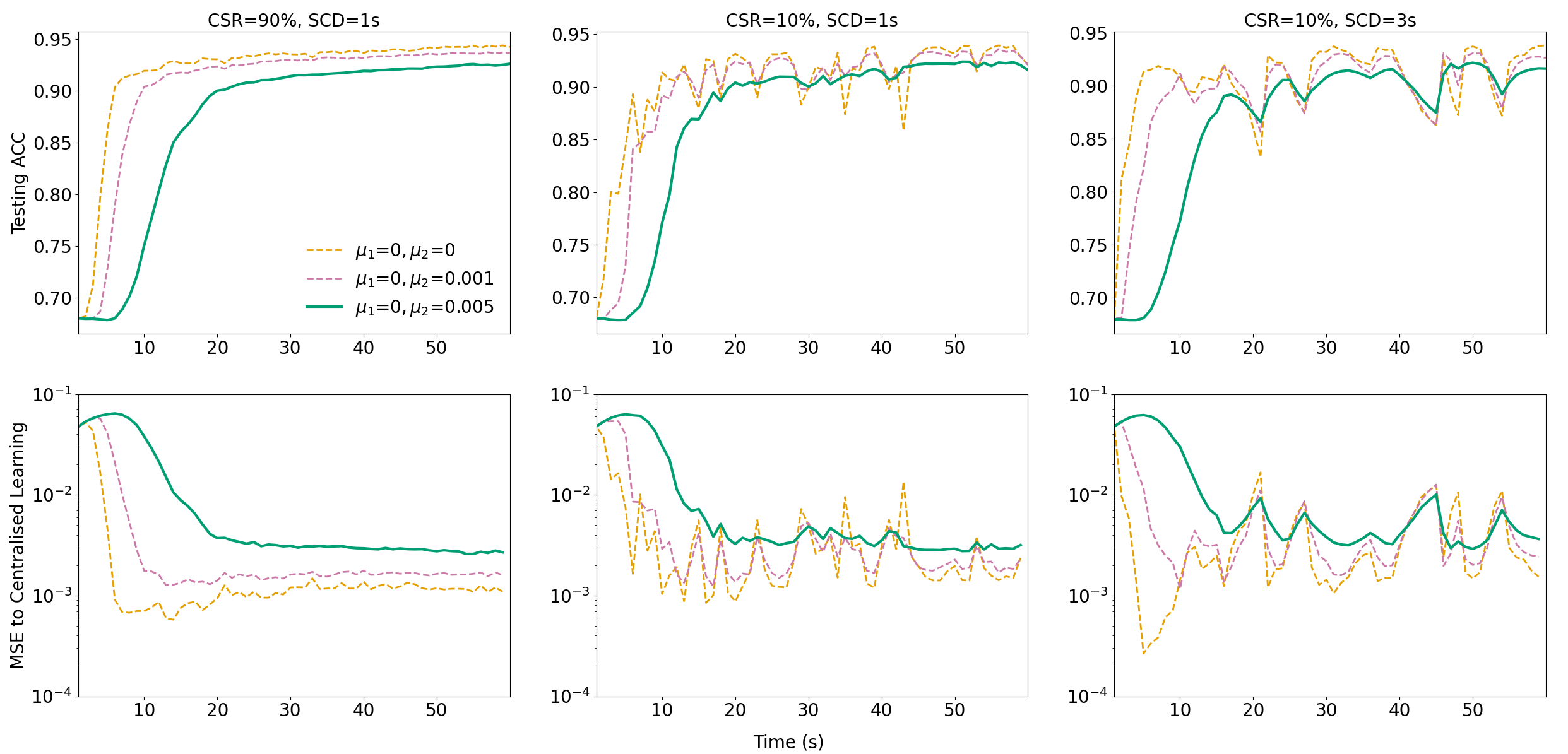}
\caption{Increasing $\mu_2$ stabilizes federated learning in scenarios with heterogeneous communication quality.}
\label{fig:Mue2}
\vspace{-0.3cm}
\end{figure*}

We set up an experiment with a DNN model with a size of $130\,kB$ trained on the MNIST~\cite{726791} dataset, which has $10$ labels as road traffic scenarios and is processed as Non-IID data partitions. Similar to the methods in~\cite{caldas2019leaf}, those partitions are divided into $110$ traffic agents, where the first 10 agents exclude a few labels and are used for pre-training as the addressed application scenarios. The pre-trained DNN model with  $68\,\%$ testing accuracy (ACC) is set as the initial model in all three layers as described in Section~\ref{sec:system}. Each of the other $100$ agents is able to train the DNN model locally and communicate to one RSU if the connection can successfully be established. As formulated in Section~\ref{sec:problem_formulation}, we vary $CSR$ and $SCD$, and consider the heterogeneous communication quality in V2X networks.

To evaluate the enhancement of the pre-trained model specifically by various $\mu_1$, we note the testing accuracy changes at time $t$ compared to the pre-trained DNN model as $\Delta ACC = ACC_t-ACC_{pre}$, and define \emph{ACC Enhancement Degree} (AED) as
\begin{equation}
    \label{Eq:AED}
    AED = (\Delta ACC^{\mu_1>0} - \Delta ACC^{\mu_1=0}) / \Delta ACC^{\mu_1=0}.
\end{equation}
The metric AED indicates the testing accuracy enhancement by increasing $\mu_1$ ($\mu_1\,>\,0$) in the set of parameters $\mathcal{M}$ (Algorithm~\ref{alg:agent}) for agents connecting to the RSUs, which can only provide the data with very limited label types. As the first column of Fig.~\ref{fig:Mue1} shows, when we set $\mu_1=0.001$ the communication quality satisfied is (CSR=100\,\%), the AED is overall positive (under $0.5\,\%$) with $\mu_2\,=\,0$ or $\mu_2\,=\,0.001$ after convergence (from around 10\,seconds). By increasing $\mu_1$, the AED is clearly raised. Each row in Fig.~\ref{fig:Mue1} shows the AED during federated learning with various $\mu_1$ and fixed $\mu_2$ with different qualities of communication. The results indicate that the AED can be obviously increased when the communication quality is rather bad (low CSR), which can oft happen in V2X networks in C-ITS due to the low priority of federated learning-related communication messages. When only 20\,\% traffic agents are successfully connected to the RSUs, $\mu_1\,=\,0.001$ can enhance the testing ACC of the federated trained DNN model up to $20\,\%$ compared to the model trained without setting $\mu_1$. 

\begin{figure*}[ht]
\centering
\begin{subfigure}[b]{0.45\textwidth}
   \includegraphics[trim=0 0 0 0 ,clip,width=1\linewidth]{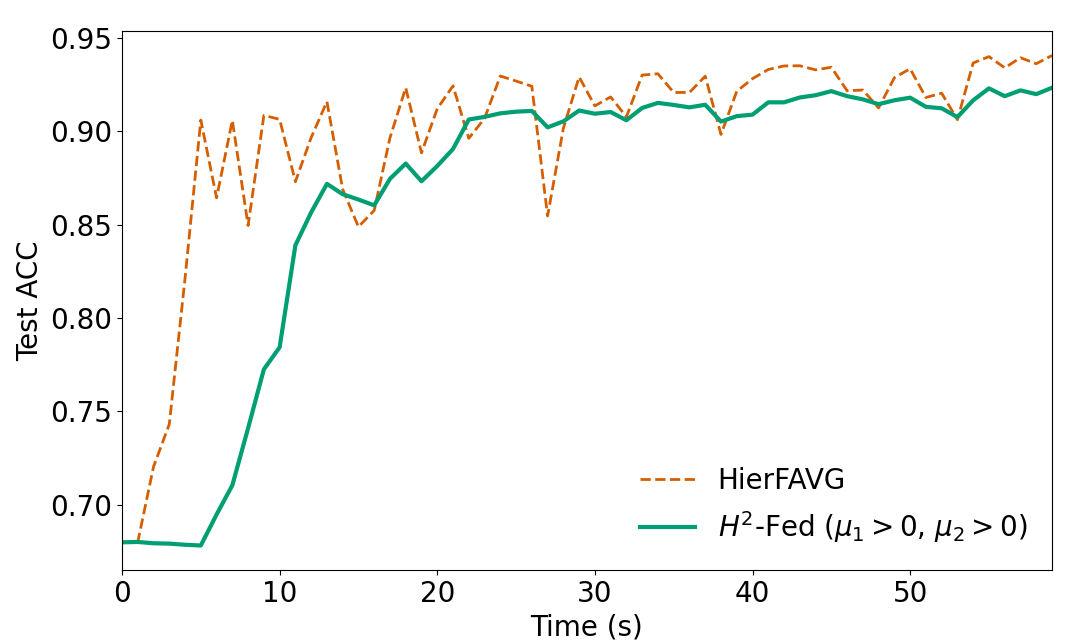}
   \caption{\centering Scenario I: Non-IID datasets only in RSU layer}
   \label{fig:scenario1} 
\end{subfigure}
\begin{subfigure}[b]{0.45\textwidth}
   \includegraphics[trim=0 0 0 0,clip,width=1\linewidth]{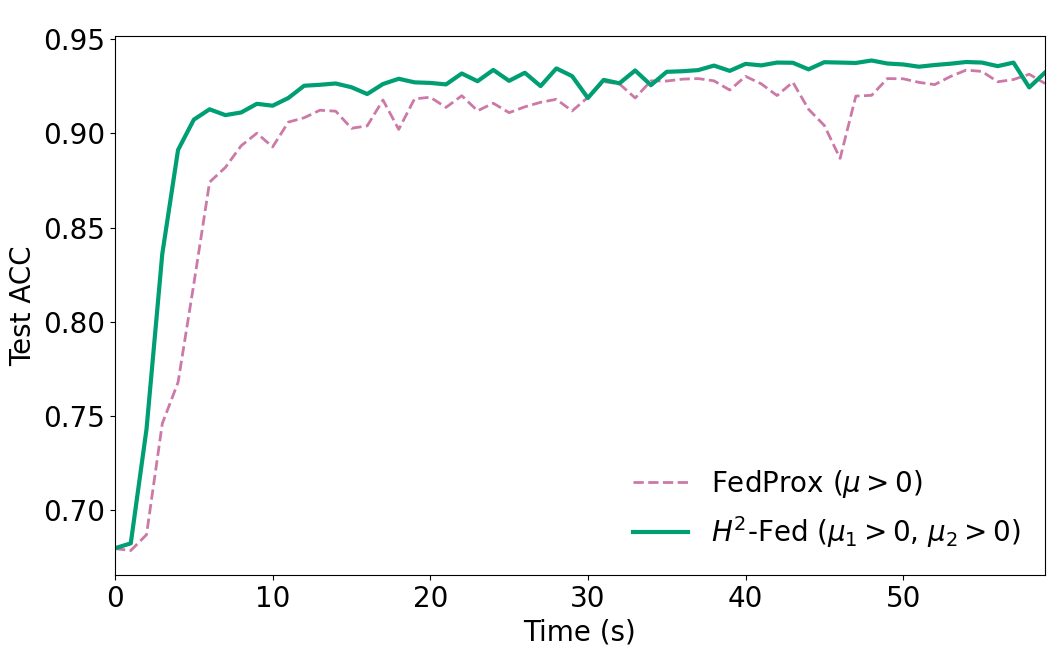}
   \caption{\centering Scenario II: Non-IID datasets only in traffic agent layer}
   \label{fig:scenario2}
\end{subfigure}
\vspace{2mm}
\caption{\centering Comparison between our framework {\myHFed} and other baseline methods for heterogeneous communication quality. Note that the baseline methods can also be seen as our framework with dedicated parameter combinations.}
\label{fig:HW}
\vspace{-0.3cm}
\end{figure*}

Next, we evaluate the effect of positive $\mu_2$ on federated learning. We notice in Fig.~\ref{fig:Mue1} that a positive value of $\mu_2$ can reduce the AED to some extent. However, for most federated learning application scenarios, a failure in the learning process is caused by unreliable connections between clients and the server. Thus, learning stability can be more significant for federated learning. The first row of Fig.~\ref{fig:Mue2} shows that the concussion in the learning process due to the low communication quality can be well stabilized through a large $\mu_2$. We also present the \emph{Mean Squared Error} (MSE) of the testing accuracy to centralized learning results in the second row of Fig.~\ref{fig:Mue2}, where the raw data from all traffic agents are collected in the cloud layer and used for centralized model training. When $\mu_2\,=\,0.005$, the jitters of the testing accuracy curve are extraordinarily coped, where the performance of federated learning is almost the same as learning with $CSR\,=\,90\,\%$. 

Finally, we compare our {\myHFed} with other two baseline methods --  \emph{FedProx}~\cite{li2020federated}, \emph{HierFAVG}~\cite{9148862} -- considering Non-IID datasets and heterogeneous communication quality with $CSR\,=\,10\,\%$ and $SCD\,=\,1\,s$. We demonstrate all those federated DNN model enhancement approaches in two different empirical C-ITS application scenarios with $100$ traffic agents and $10$ RSUs. In \emph{Scenario I}, the datasets across RSUs are Non-IID, while the datasets in all traffic agents under one RSU have the same distribution. In contrast, the datasets across traffic agents at one RSU are Non-IID in \emph{Scenario II}, where each RSUs has the same data distribution. 

Fig.~\ref{fig:scenario1} shows our framework {\myHFed} can enhance the pre-trained model stably from beginning to convergence, while the jitters of \emph{HierFAVG} curve are more visibly affected by the bad communication situation. Furthermore, as shown in Fig.~\ref{fig:scenario2}, our framework {\myHFed} in \emph{Scenario II} outperforms \emph{FedProx} remarkably, as all models at each RSU are pre-aggregated multiple times, which accelerates the convergence of federated learning. The comparison results indicate that our framework can address hierarchical heterogeneity by adapting more tunable parameters which can result in the success of federated learning in C-ITS. 

\section{Conclusion}
\label{sec:conclusion}
In this work, we highlight one of the most important bottlenecks in AI model training for C-ITS -- insufficient data, which can be solved by federated learning in V2X communication networks. We address the individual heterogeneity problems in different layers of C-ITS and proposed a novel method, as well as an associated general framework named {\myHFed} for enhancing the pre-trained DNN model. Our empirical evaluation demonstrates that the framework can guarantee the success of federated learning even with large-scale disconnected traffic agents in C-ITS. The comparison with other baseline methods indicates the flexibility and generality of {\myHFed} that provides opportunities to well balance the trade-off between accuracy and stability in different application scenarios. 
Besides, we believe that in our framework the robustness of federated learning can be further improved by dynamic parameter settings, which will be validated in future simulations. Moreover, time-variant latency due to road traffic topology and efficient encoding for dissemination of large DNN models remain.

\IEEEtriggeratref{13} 
\bibliography{ref}
\bibliographystyle{IEEEtran}

\end{document}